\documentclass[11pt,a4paper,table,xcdraw]{article}
\usepackage[hyperref]{emnlp-ijcnlp-2019}
\usepackage{times}
\usepackage{latexsym}

\usepackage{graphicx}
\usepackage{caption} 
\usepackage{subcaption}
\usepackage{stfloats}

\usepackage{multirow}
\usepackage{xcolor}

\usepackage{algorithm}
\usepackage[noend]{algpseudocode}

\usepackage{url}
\usepackage{appendix}

\aclfinalcopy 

\usepackage{fancyhdr}
\title{Improving Quality and Efficiency \\in Plan-based Neural Data-to-Text Generation}

\author{Amit Moryossef$^\dagger \;\;\;$  Ido Dagan$^\dagger \;\;\;$ Yoav Goldberg$^\dagger$$^\ddagger$ \\
\texttt{amitmoryossef@gmail.com, \{dagan,yogo\}@cs.biu.ac.il} \\
\\ $^\dagger$Bar Ilan University, Ramat Gan, Israel \\
$^\ddagger$Allen Institute for Artificial Intelligence}

\date{}

\begin{document}
\maketitle
\begin{abstract}
We follow the step-by-step approach to neural data-to-text generation we proposed in \citet{step-by-step}, in which the generation process is divided into a text-planning stage followed by a plan-realization stage.
We suggest four extensions to that framework: (1) we introduce a trainable neural planning component that can generate effective plans several orders of magnitude faster than the original planner; (2) we incorporate typing hints that improve the model's ability to deal with unseen relations and entities; (3) we introduce a verification-by-reranking stage that substantially improves the faithfulness of the resulting texts; (4) we incorporate a simple but effective referring expression generation module. These extensions result in a generation process that is faster, more fluent, and more accurate. \end{abstract}

\thispagestyle{fancy}

\section{Introduction}
In the data-to-text generation task (D2T), the input is data encoding facts (e.g., a table, a set of tuples, or a small knowledge graph), and the output is a natural language text representing those facts.\footnote{In this paper, we focus on a setup where the desired output represents \emph{all} and \emph{only} the facts expressed in the dataset. Other variants also involve content selection, allowing the process to select which subset of the facts to express.} In neural D2T, the common approaches train a neural end-to-end encoder-decoder system that encodes the input data and decodes an output text. In recent work \cite{step-by-step} we proposed to adopt ideas from ``traditional'' language generation approaches (i.e. \citet{reiter2000building, walker2007individual, DBLP:journals/corr/GattK17}) that separate the generation into a \emph{planning} stage that determines the order and structure of the expressed facts, and a \emph{realization} stage that maps the plan to natural language text. We show that by breaking the task this way, one can achieve the same fluency of neural generation systems while being able to better control the form of the generated text and to improve its correctness by reducing missing facts and ``hallucinations'', common in neural systems.

In this work we adopt the step-by-step framework of \citet{step-by-step} and propose four independent extensions that improve aspects of our original system: we suggest a new plan generation mechanism, based on a trainable-yet-verifiable neural decoder, that is orders of magnitude faster than the original one (\S\ref{sec:planner}); we use knowledge of the plan structure to add typing information to plan elements. This improves the system's performance on unseen relations and entities (\S\ref{sec:types});  the separation of planning from realizations allows the incorporation of a simple output verification heuristic that drastically improves the correctness of the output (\S\ref{sec:rerank}); and finally we incorporate a post-processing referring expression generation (REG) component, as proposed but not implemented in our previous work, to improve the naturalness of the resulting output (\S\ref{sec:reg}).

\section{Step-by-step Generation}
We provide a brief overview of the step-by-step system. See \citet{step-by-step} for further details.
The system works in two stages. The first stage (planning) maps the input facts (encoded as a directed, labeled graph, where nodes represent entities and edges represent relations) to text plans, while the second stage (realization) maps the text plans to natural language text. 

The text plans are a sequence of sentence plans---each of which is a tree--- 
representing the ordering of facts and entities within the sentence. In other words, the plans determine the separation of facts into sentences, the ordering of sentences, and the ordering of facts and entities within each sentence. This stage is \emph{completely verifiable}: the text plans are guaranteed to faithfully encode all and only the facts from the input.
The realization stage then translates the plans into natural language sentences, using a neural sequence-to-sequence system, resulting in fluent output. 

\section{Fast and Verifiable Planner}
\label{sec:planner}
The data-to-plan component in \citet{step-by-step} exhaustively generates all possible plans, scores them using a heuristic, and chooses the highest scoring one for realization. 
While this is feasible with the small input graphs in the WebNLG challenge \cite{colin2016webnlg}, it is also very computationally intensive, growing exponentially with the input size. We propose an alternative planner which works in linear time in the size of the graph and remains verifiable: generated plans are guaranteed to represent the input faithfully.

The original planner works by first enumerating over all possible splits into sentences (sub-graphs), and for each sub-graph enumerating over all possible undirected, unordered, Depth First Search (DFS) traversals, where each traversal corresponds to a sentence plan. Our planner combines these into a single process.
It works by performing a series of what we call \emph{random truncated DFS traversals}. In a DFS traversal, a node is visited, then its children are visited recursively in order. Once all children are visited, the node ``pops'' back to the parent. In a \emph{random truncated traversal}, the choice of which children to visit next, as well as whether to go to the next children or to ``pop'', is non-deterministic (in practice, our planner decides by using a neural-network controller). Popping at a node before visiting all its children truncates the DFS: further descendants of that node will not be visited in this traversal. It behaves as a DFS on a graph where edges to these descendants do not exist. Popping the starting node terminates the traversal.

Our planner works by choosing a node with a non-zero degree and performing a truncated DFS traversal from that node. Then, all edges visited in the traversal are removed from the input graph, and the process repeats (performing another truncated DFS) until no more edges remain.  Each truncated DFS traversal corresponds to a sentence plan, following the DFS-to-plan procedure of \citet{step-by-step}: the linearized plan is generated incrementally at each step of the traversal. This process is linear in the number of edges in the graph.

At training time, we use the plan-to-DFS mapping to perform the correct sequence of traversals, and train a neural classifier to act as a controller, choosing which action to perform at each step. At test time, we use the controller to guide the truncated DFS process. This mechanism is inspired by transition based parsing \cite{nivre2008integrating}. The action set at each stage is dynamic. During traversal, it includes the available children at each stage and \textsc{pop}. Before traversals, it includes a \emph{choose-i} action for each available node $n_i$. We assign a score to each action, normalize with softmax, and train to choose the desired one using cross-entropy loss. At test time, we either greedily choose the best action, or we can sample plans by sampling actions according to their assigned probabilities.

\noindent\textbf{Feature Representation and action scoring.}
Each graph node $n_i$ corresponds to an entity $x_{n_i}$, and has an associated embedding vector $\mathbf{x_{n_i}}$. Each relation $r_i$ is associated with an embedding vector $\mathbf{r_i}$. Each labeled input graph edge $e_k = (n_i, r_\ell, n_j)$ is represented as a projected concatenated vector $\mathbf{e_k}=\mathbf{E}(\mathbf{x_{n_i}};\mathbf{r_\ell};\mathbf{x_{n_j}})$, where $\mathbf{E}$ is a projection matrix. Finally,   each node $n_i$ is then represented as a vector $\mathbf{n_i} = \mathbf{V}[\mathbf{x_{n_i}};\sum_{e_j\in \pi(i)}\mathbf{e_j};\sum_{e_j\in\pi^{-1}(i)}\mathbf{e_j}]$, where $\pi(i)$ and $\pi^{-1}(i)$ are the incoming and outgoing edges from node $n_i$. 
The traverse-to-child-via-edge-$e_j$ action is represented as $\mathbf{e_j}$, choose-node-i is represented as $\mathbf{n_i}$ and pop-to-node-i is represented as $\mathbf{n_i}+\mathbf{p}$ where $\mathbf{p}$ is a learned vector. The score for an action $a$ at time $t$ is calculated as a dot-product between the action representation and the LSTM state over the symbols generated in the plan so far. Thus, each decision takes into account the immediate surrounding of the node in the graph, and the plan structure generated so far.

\noindent\textbf{Speed} On a 7 edges graph, the planner of \citet{step-by-step} takes an average of 250 seconds 
to generate a plan, while our planner takes 0.0025 seconds, 5 orders of magnitude faster.

\section{Incorporating typing information for unseen entities and relations}
\label{sec:types}
In \citet{step-by-step}, the sentence plan trees were linearized into strings that were then fed to a neural machine translation decoder (OpenNMT) \cite{klein2017opennmt} with a copy mechanism. This linearization process is lossy, in the sense that the linearized strings do not explicitly distinguish between symbols that represent \emph{entities} (e.g., \texttt{BARACK\_OBAMA}) and symbols that represent \emph{relations} (e.g., \texttt{works-for}). While this information can be deduced from the position of the symbol within the structure, there is a benefit in making it more explicit. In particular, the decoder needs to act differently when decoding relations and entities: entities are copied, while relations need to be verbalized. By making the typing information explicit to the decoder, we make it easier for it to generalize this behavior distinction and apply it also for \emph{unseen} entities and relations.  We thus expect the typing information to be especially useful for the unseen part of the evaluation set.

We incorporate typing information by concatenating to the embedding vector of each input symbol one of three embedding vectors, \textbf{S}, \textbf{E} or \textbf{R}, where \textbf{S} is concatenated to structural elements (opening and closing brackets), \textbf{E} to entity symbols and \textbf{R} to relation symbols. 

\section{Output verification}
\label{sec:rerank}
While the plan generation stage is guaranteed to be faithful to the input, the translation process from plans to text is based on a neural seq2seq model and may suffer from known issues with such models: hallucinating facts that do not exist in the input, repeating facts, or dropping facts. While the clear mapping between plans and text helps to reduce these issues greatly, the system in \citet{step-by-step} still has 2\% errors of these kinds.

\paragraph{Existing approaches: soft encouragement via neural modules.} Recent work in neural text generation and summarization attempt to address these issues by trying to map the textual outputs back to structured predicates, and comparing these predicates to the input data. \citet{kiddon2016globally} uses a neural checklist model to avoid the repetition of facts and improve coverage. \citet{agarwal2018char2char} generate $k$-best output candidates with beam search, and then try to map each candidate output back to the input structure using a reverse seq2seq model trained on the same data. They then select  the highest scoring output candidate that best translates back to the input. \citet{mohiuddin2019revisiting} reconstructs the input in training time, by jointly learning a back-translation model and enforcing the back-translation to reconstruct the input. 
Both of these approaches are ``soft'' in the sense that they crucially rely on the internal dynamics or on the output of a neural network module that may or may not be correct.

\paragraph{Our proposal: explicit verification.} The separation between planning and realization provided by the step-by-step framework allows incorporating a robust and straightforward \emph{verification step}, that does not rely on brittle information extraction procedures or trust neural network models. 

The plan-to-text generation handles each sentence individually and translates entities as copy operations. We thus have complete knowledge of the generated entities and their locations. We can then assess the correctness of an output sentence by comparing\footnote{We use 
Levenshtein-distance \cite{levenshtein1966binary}.} its sequence of entities to the entity sequence in the corresponding sentence plan, which is guaranteed to be complete.

We then decode $k$-best outputs and rerank them based on their correctness scores, tie-breaking using model scores. We found empirically that, with a beam of size 5 we find at least one candidate with an exact match to the plan's entity sequence in 99.82\% of the cases for seen entities and relations compared to 98.48\% at 1-best, and 72.3\% for cases of unseen entities and relations compared to 58.06\% at 1-best. In the remaining cases, we set the system to continue searching by trying other plans, by going down the list of plans (when using the exhaustive planner of \citet{step-by-step}) or by sampling a new plan (when using the linear time planner suggested in this paper).


\section{Referring Expressions}
\label{sec:reg}
The step-by-step system generates entities by first generating an indexed entity symbols, and then lexicalizing each symbol to the string associated with this entity in the input structure (i.e., all occurrences of the entity \emph{11TH MISSISSIPPI INFANTRY MONUMENT} will be lexicalized with the full name rather than ``\emph{it}'' or ``\emph{the monument}''). This results in correct but somewhat unnatural structures. In contrast, end-to-end neural generation systems are trained on text that includes referring expressions, and generate them naturally as part of the decoding process, resulting in natural looking text. However, the generated referring expressions are sometimes incorrect. \citet{step-by-step} suggests the possibility of handling this with a post-processing referring-expression generation step (REG). Here, we propose a concrete REG module and demonstrate its effectiveness. One option is to use a supervised REG module \cite{ferreira2018neuralreg}, that is trained to lexicalize in-context mentions. Such an approach is sub-optimal for our setup as it is restricted to the entities and contexts it seen in training, and is prone to error on unseen entities and contexts.

Our REG solution lexicalizes the first mention of each entity as its associated string and attempts to generate referring expressions to subsequent mentions. The generated referring expressions can take the form ``\textsc{Pron}'', ``\textsc{X}'' or ``\textsc{the X}'' where \textsc{Pron} is a pronoun\footnote{One of \emph{he, his, him, himself, she, her, hers, herself, they, them, theirs, it, its, itself.}}, and \textsc{X} is a word appearing in the entity's string (allowing, e.g., \emph{John}, or \emph{the monument}).
We also allow referring to its entity with its entire associated string. We restrict the set of allowed pronouns for each entity according to its type (male, female, plural-animate, unknown-animate, inanimate).\footnote{We extract the types from DBPedia pages for the entities. In case we cannot deduce a type, we do not allow any pronoun.} We then take, for each entity mention individually, the referring expression that receives the best language model score in context, using a strong unsupervised neural LM (BERT \cite{devlin2018bert}). The system is guaranteed to be correct in the sense that it will not generate wrong pronouns. It also has failure modes: it is possible for the system to generate ambiguous referring expressions (e.g., \emph{John is Bob's father. \underline{He} works as a nurse.}), and may lexicalize \emph{Boston University} as \emph{Boston}. We find that the second kind of mistake is rare as it is handled well by the language model. It can also be controlled by manually restricting the set of possible referring expression to each entity. Similarly, it is easy to extend the system to support other lexicalizations of entities by extending the sets of allowed lexicalizations (for example, supporting abbreviations, initials or nicknames) either as user-supplied inputs or using heuristics.



\section{Evaluation and Results}
We evaluate each of the introduced components separately. Tables listing their interactions are available in the appendix. The appendix also lists some qualitative outputs.
The main trends that we observe are:
\begin{itemize}
    \item The new planner causes a small drop in BLEU, but is orders of magnitude faster (\S\ref{eval:planner}).
    \item Typing information causes a negligible drop in BLEU overall, but improves results substantially for the \emph{unseen} portion of the dataset (\S\ref{eval:typing}).
    \item The verification step is effective at improving the faithfulness of the output, practically eliminating omitted and overgenerated facts, reducing the number of wrong facts, and increasing the number of correctly expressed facts. This is based on both manual and automatic evaluations. (\S\ref{eval:verification}).
    \item The referring expression module is effective, with an intrinsic correctness of 92.2\%. It substantially improves BLEU scores. (\S\ref{eval:reg}).
\end{itemize}

\paragraph{Setup} We evaluate on the WebNLG dataset \cite{colin2016webnlg}, comparing to the step-by-step systems described in \citet{step-by-step}, which are state of the art. Due to randomness inherent in neural training, our reported automatic evaluation measures are based on an average of 5 training runs of each system (neural planner and neural realizer), each run with a different random seed. 


\subsection{Neural Planner vs Exhaustive Planner}\label{eval:planner}
\begin{figure*}[t]
    \begin{center}
    \begin{subfigure}{.44\linewidth}
        \includegraphics[width=\linewidth]{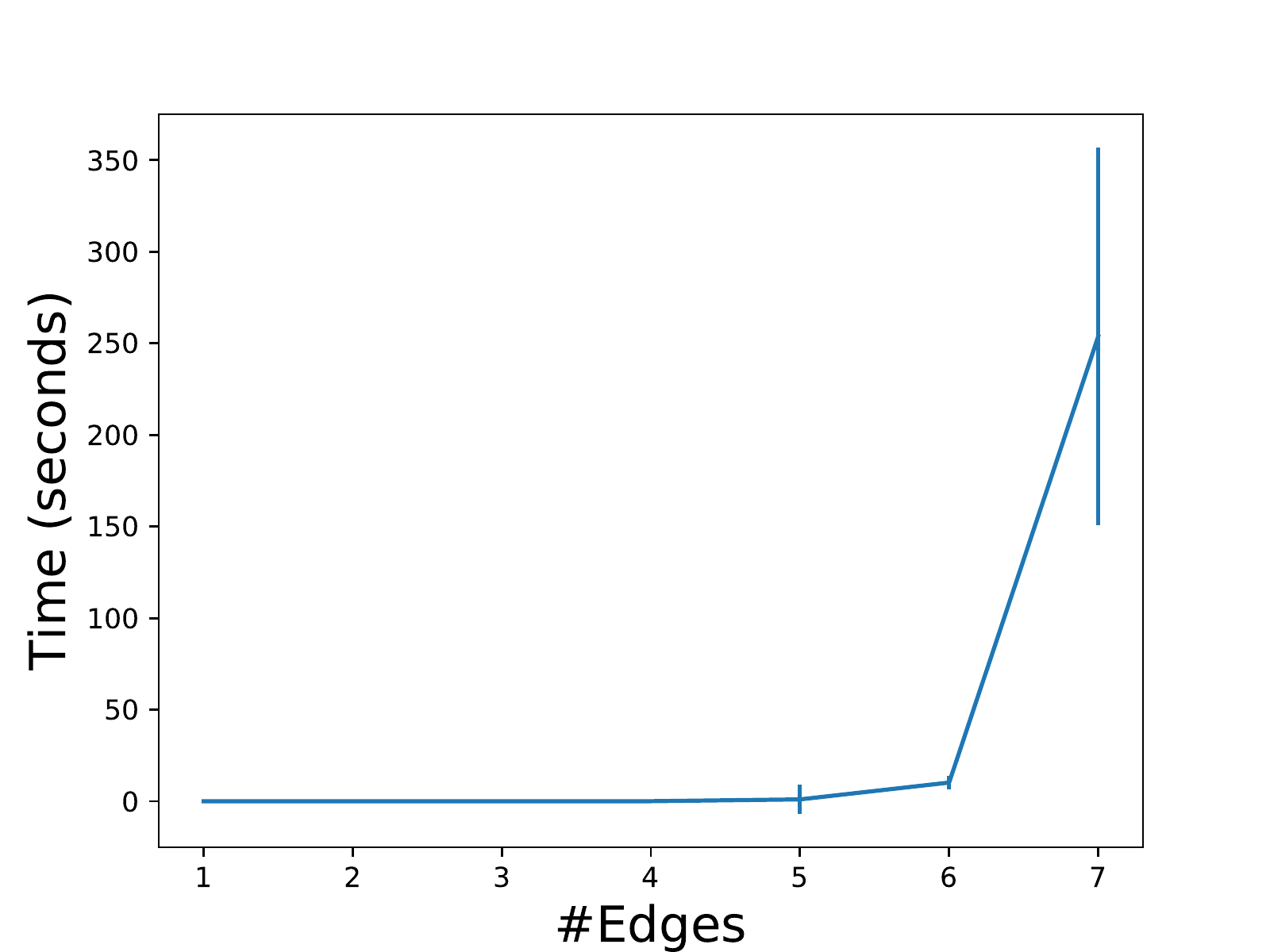}
        \caption{Exhaustive Planner}
        \label{fig:timing-naive}
    \end{subfigure}
    \begin{subfigure}{.44\linewidth}
        \includegraphics[width=\linewidth]{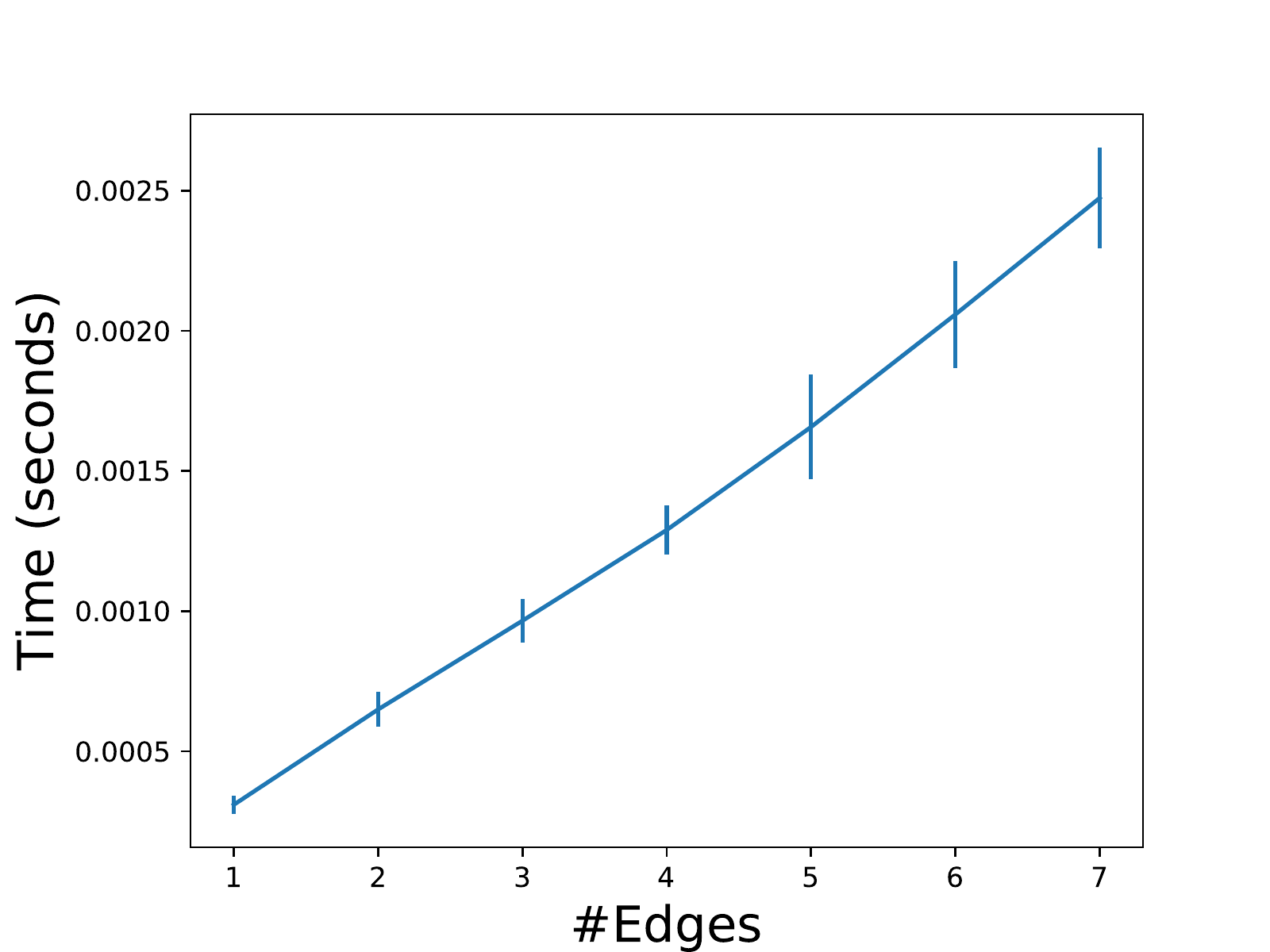}
        \caption{Neural Planner}
        \label{fig:timing-neural}
    \end{subfigure}
    \end{center}
    \caption{Average (+std) planning time (seconds) for different graph sizes, for exhaustive vs neural planner.}
    \label{fig:timing}
\end{figure*}

We compare the exhaustive planner from \citet{step-by-step} to our neural planner, by replacing the planner component in the \citet{step-by-step} system. Moving to the neural planner exhibits a small drop in BLEU (46.882 dropped to 46.506). However, figure \ref{fig:timing} indicates 5 orders of magnitude (100,000x) speedup for graphs with 7 edges, and a linear growth in time for number of edges compared to exponential time for the exhaustive planner.

\subsection{Effect of Type Information}\label{eval:typing}
We repeat the coverage experiment in \cite{step-by-step}, counting the number of output texts that contain all the entities in the input graph, and, of these text, counting the ones in which the entities appear in the exact same order as the plan. 
Incorporating typing information reduced the number of texts not containing all entities by 18\% for the \emph{seen} part of the test set, and 16\% for the \emph{unseen} part. Moreover, for the text containing all entities, the number of texts that did not follow the plan's entity order is reduced by 46\% for the \emph{seen} part of the test set, and by 35\% for the \emph{unseen} part. We also observe a small drop in BLEU scores, which we attribute to some relations being verbalized more freely (though correctly).
\begin{table}[t]

\begin{center}
\resizebox{\linewidth}{!}
{
\begin{tabular}{|l|l|l||l|l|}
\hline
 & \textbf{Moryosef et al} & \textbf{Moryosef et al} & \textbf{Exhaustive} & \textbf{Neural} \\
 & \textbf{StrongNeural} & \textbf{BestPlan} & \textbf{+Verify} & \textbf{+Verify} \\ \hline
\textbf{Expressed} & 360 & 417 & \textbf{426} & 405 \\ \hline
\textbf{Omitted} & 41 & 6 & \textbf{0} & 2 \\ \hline
\textbf{Wrong} & 39 & 17 & \textbf{14} & 30 \\ \hline
\textbf{Over-generation} & 29 & 3 & \textbf{0} & 4 \\ \hline
\textbf{Wrong REG} & - & - & \textbf{0} & 3 \\ \hline
\end{tabular}
}
\end{center}
\caption{Manual correctness analysis comparing our systems with the ones from \citet{step-by-step}.}
\label{table:manual}
\end{table}

\subsection{Effect of Output Verification}\label{eval:verification}
The addition of output verification resulted in negligible changes in BLEU, reinforcing that automatic metrics are not sensitive enough to output accuracy. We thus performed manual analysis, following the procedure in \citet{step-by-step}. We manually inspect 148 samples from the seen part of the test set, containing 440 relations, counting expressed, omitted, wrong and over-generated (hallucinated) facts.\footnote{A wrong fact is one in which a fact exists between the two entities, but the text implies a different fact from the graph, while over-generated is either repeating facts or inventing new facts.} We compare to the \textbf{StrongNeural} and \textbf{BestPlan} systems from \citet{step-by-step}. Results in Table \ref{table:manual} indicate that the effectiveness of the verification process in ensuring correct output, reducing the already small number of ommited and overgenerated facts to 0 (with the exhaustive planner) and keeping it small (with the fast neural planner).

\subsection{Referring Expression Module}\label{eval:reg}

\paragraph{Intrinsic evaluation of the REG module.}
We manually reviewed 1,177 pairs of entities and referring expressions generated by the system.
We find that 92.2\% of the generated referring expressions refer to the correct entity.

From the generated expressions, 325 (27.6\%) were pronouns, 192 (16.3\%) are repeating a one-token entity as is, and 505 (42.9\%) are generating correct shortening of a long entity. In 63 (5.6\%) of the cases the system did not find a good substitute and kept the entire entity intact. Finally, 92 (7.82\%) are wrong referrals. Overall, 73.3\% of the non-first mentions of entities were replaced with suitable shorter and more fluent expressions.

\paragraph{Effect on BLEU scores.}
As can be seen in Table \ref{table:bleu-reg}, using the REG module increases BLEU scores for both the exhaustive and the neural planner.
\begin{table}[h]
\begin{center}
\begin{tabular}{l|l|l|}
\cline{2-3}
 & \textbf{-} & \textbf{REG} \\ \hline
\multicolumn{1}{|l|}{\textbf{Exhaustive Planner}} & 46.882 & 47.338 \\ \hline
\multicolumn{1}{|l|}{\textbf{Neural Planner}} & 46.506 & 47.124 \\ \hline
\end{tabular}
\end{center}
\caption{Effect of the REG component on BLEU score}
\label{table:bleu-reg}
\end{table}

\section{Conclusions}
We adopt the planning-based neural generation framework of \citet{step-by-step} and extend it to be orders of magnitude faster and produce more correct and more fluent text. We conclude that these extensions not only improve the system of \citet{step-by-step} but also highlight the flexibility and advantages of the step-by-step framework for text generation. 

\clearpage

\section*{Acknowledgements}
This work was supported in part by the German Research Foundation through the German-Israeli Project Cooperation (DIP, grant DA 1600/1-1) and by a grant from Reverso and Theo Hoffenberg.

\bibliography{emnlp-ijcnlp-2019}
\bibliographystyle{acl_natbib}
\clearpage

\begin{table*}[b]
\begin{center}
\begin{tabular}{ll|c|c|c|c|}
\cline{3-6}
  &                     & \multicolumn{2}{l|}{\textbf{Exhaustive Planning}}                        & \multicolumn{2}{l|}{\textbf{Neural Planning}}                         \\ \cline{3-6} 
  &                     & \multicolumn{1}{l|}{\textbf{-}} & \multicolumn{1}{l|}{\textbf{REG}} & \multicolumn{1}{l|}{\textbf{-}} & \multicolumn{1}{l|}{\textbf{REG}} \\ \hline
\multicolumn{1}{|l|}{}                                        & \textbf{No Verification}     & \cellcolor[HTML]{D4CE70}46.882  & \cellcolor[HTML]{64BF86}47.338    & \cellcolor[HTML]{F7C266}46.506  & \cellcolor[HTML]{ABC973}47.124      \\ \cline{2-6} 
\multicolumn{1}{|l|}{\multirow{-2}{*}{\textbf{No types}}}          & \textbf{Verified Output} & \cellcolor[HTML]{D1CF67}46.896  & \cellcolor[HTML]{56BD85}47.392    & \cellcolor[HTML]{FDB070}46.412  & \cellcolor[HTML]{75C083}47.05       \\ \hline
\multicolumn{1}{|l|}{}                                        & \textbf{No Verification}     & \cellcolor[HTML]{E79072}46.194  & \cellcolor[HTML]{EDD56A}46.768    & \cellcolor[HTML]{E47D75}45.902  & \cellcolor[HTML]{FBCD6B}46.628      \\ \cline{2-6} 
\multicolumn{1}{|l|}{\multirow{-2}{*}{\textbf{With Typing}}} & \textbf{Verified Output} & \cellcolor[HTML]{E67C73}46.072  & \cellcolor[HTML]{FCCC6A}46.614    & \cellcolor[HTML]{EC9B6E}46.166  & \cellcolor[HTML]{DDD270}46.834      \\ \hline
\end{tabular}
\end{center}
\caption{Average BLEU score for every combination of methods (avg of 5 independent runs).}
\label{table:bleu}
\end{table*}
\appendix
\appendixpage
\addappheadtotoc

We report some additional results.

\section{Interaction of different components}
We introduced 4 components: neural planner instead of exhaustive one, adding type information, adding output verification stage, and incorporating a referring expression generation (REG). In Table \ref{table:bleu} we report BLEU scores 
 \cite{papineni2002bleu} for all 16 combinations of components. The numbers are averages of 5 runs with different random seeds.


\section{REG Error Analysis}
We perform further analysis of the errors of the unsupervised LM based REG module. We categorise all entities into 3 groups: (1) names of people; (2) locations (cities / counties / countries); and (3) places and objects.

For person names, the module did not produce any errors, selecting either a correct pronoun, or either the first or last name of a person, all valid refferences.

For location names, we observe two distinct error types, both relating to our module's restriction to predict a single MASK token. The first type is in cases like ``city, country'' or ``county, country'', where the more specific location is not in the LM vocabulary, and cannot be predicted with a single token. For example, in ``Punjab, Pakistan'', Punjab is not contained in the vocabulary as a single token, causing the model to select ``Pakistan'', which we consider a mistake. The second type is when a city name is longer than a single token, as in ``New York''. While it is common to refer to ``New Jersey'' as ``Jersey'', it is wrong to refer to ``New York'' as either ``New'' or ``York'', and as BERT can only fill in one MASK token, it chooses only one (in this case ``York'').

Finally, for places and objects, we also identify to mistake types. The first occurs for multi-token entities. While for some cases it is possible to select the correct one (i.e., ``Agra Airport'' $\rightarrow$ ``The Airport'' or ``Boston University'' $\rightarrow$ ``The University''), in other cases it is not possible (i.e., ``Baked Alaska'', where choosing either word does not produce a useful reference). 
The second type occurs with names of objects, like books titles. For example, for the entity ``A Severed Wasp'' we would like the model to predict ``The Book''. However, as we only allow either pronouns or words from the original entity, the model cannot produce ``The book'', producing the erroneous ``The Wasp'' instead.

\section{Output Examples}
The following output examples demonstrate the kinds of texts produces by the final system. 
The following outputs are correct, expressing all and only the facts from their input graphs. We enumerate them as number of facts:
\begin{enumerate}
    \item The leader of \textbf{Azerbaijan} is \textbf{Artur Rasizade}.
    \item \textbf{Baked Alaska}, containing \textbf{Sponge Cake}, is from \textbf{France}.
    \item \textbf{Above The Veil}, written by \textbf{Garth Nix}, is available in \textbf{Hardcover} and has \textbf{248} pages.
    \item The \textbf{Akita Museum Of Art} is located in \textbf{Japan} where the \textbf{Brazilians In Japan} are an ethnic group. \textbf{The Museum} is located in \textbf{Akita, Akita} which is part of \textbf{Akita Prefecture} .
    \item The \textbf{AWH Engineering College} in \textbf{Kuttikkattoor}, \textbf{Kerala} has \textbf{Mahé, India} to its northwest . \textbf{The College} was established in \textbf{2001} and has a staff of \textbf{250}.
\end{enumerate}

An example where the system failed, producing a wrong lexicalization of a fact is: ``The \textbf{AWH Engineering College} is located in the state of \textbf{Kerala}, \textbf{Kochi}, in \textbf{India}. The largest city in \textbf{India} is \textbf{Mumbai} and the river is the \textbf{Ganges}''. In this example, the input entity \textbf{Kochi} refers to the leader of \textbf{Kerala}, and not tpo the location (although there is also a location by that name).  The text lexicalizes this fact such that \textbf{Kerala} and \textbf{Kochi} are related, but with a relation of \emph{part-of}, implying \textbf{Kerala} is in \textbf{Kochi}.

\end{document}